\documentclass[letterpaper, 10 pt, conference]{ieeeconf}  

\IEEEoverridecommandlockouts                              

\usepackage{amsmath} 
\usepackage{amssymb}  
\usepackage{graphicx}
\usepackage{tabularx}
\usepackage{float}
\usepackage{wrapfig}
\usepackage{multirow}
\usepackage{hyperref}
\usepackage{xcolor}

\usepackage{booktabs}

\title{\LARGE \bf
Physics-Informed Neural Controlled Differential Equations for Scalable Long Horizon Multi-Agent Motion Forecasting
}

\author{Shounak Sural$^{1*}$, Charles Kekeh$^{2}$, Wenliang Liu$^{2}$, Federico Pecora$^{2}$ and Mouhacine Benosman$^{2}$
\thanks{*Work done during internship at Amazon Robotics}
\thanks{$^{1}$Carnegie Mellon University, Pittsburgh, PA, USA
        {\tt\small ssural@andrew.cmu.edu}}%
\thanks{$^{2}$Amazon Robotics, North Reading, MA, USA
        {\tt\small \{liuwll,ckkekeh,fpecora,mbenos\}@amazon.com}}%
}

\begin{document}

\maketitle
\thispagestyle{empty}
\pagestyle{empty}

\begin{abstract}
Long-horizon motion forecasting for multiple autonomous robots is challenging due to non-linear agent interactions, compounding prediction errors, and continuous-time evolution of dynamics. Learned dynamics of such a system can be useful in various applications such as travel time prediction, prediction-guided planning and generative simulation. In this work, we aim to develop an efficient trajectory forecasting model conditioned on multi-agent goals. Motivated by the recent success of physics-guided deep learning for partially known dynamical systems, we develop a model based on neural Controlled Differential Equations (CDEs) for long-horizon motion forecasting. Unlike discrete-time methods such as RNNs and transformers, neural CDEs operate in continuous time, allowing us to combine physics-informed constraints and biases to jointly model multi-robot dynamics. Our approach, named PINCoDE (Physics-Informed Neural Controlled Differential Equations), learns differential equation parameters that can be used to predict the trajectories of a multi-agent system starting from an initial condition. PINCoDE is conditioned on future goals and enforces physics constraints for robot motion over extended periods of time. We adopt a strategy that scales our model from 10 robots to 100 robots without the need for additional model parameters, while producing predictions with an average ADE below 0.5 m for a 1-minute horizon. Furthermore, progressive training with curriculum learning for our PINCoDE model results in a $2.7\times$ reduction of forecasted pose error over 4 minute horizons compared to analytical models.

\end{abstract}

\section{Introduction}
Spatio-temporal dynamics modeling for multiple interacting autonomous mobile robots (AMRs) can be a challenging task \cite{baniodeh2025scalinglawsmotionforecasting, wayformer, trajectron++, motionlm}. Non-linear coupled dynamics of such robots, hidden agents and irregular sampling times across multiple robots often add to the complexity. Such dynamics modeling can benefit from methods that learn the continuous-time evolution of the joint state of the multi-agent system. Modern robot motion forecasting methods primarily rely on transformers or GNNs to model robot interactions and then predict robot poses in discrete-time with auto-regressive models \cite{trajectron++,wayformer,motionlm}. However, many such auto-regressive methods tend to predict divergent trajectories with errors compounding over time steps, especially in scenarios such as sharp turns, sudden events and cases with missing or irregularly-timestamped data \cite{compounding_error}. 

\begin{figure}[t]
    \centering
    \includegraphics[width=0.9\linewidth]{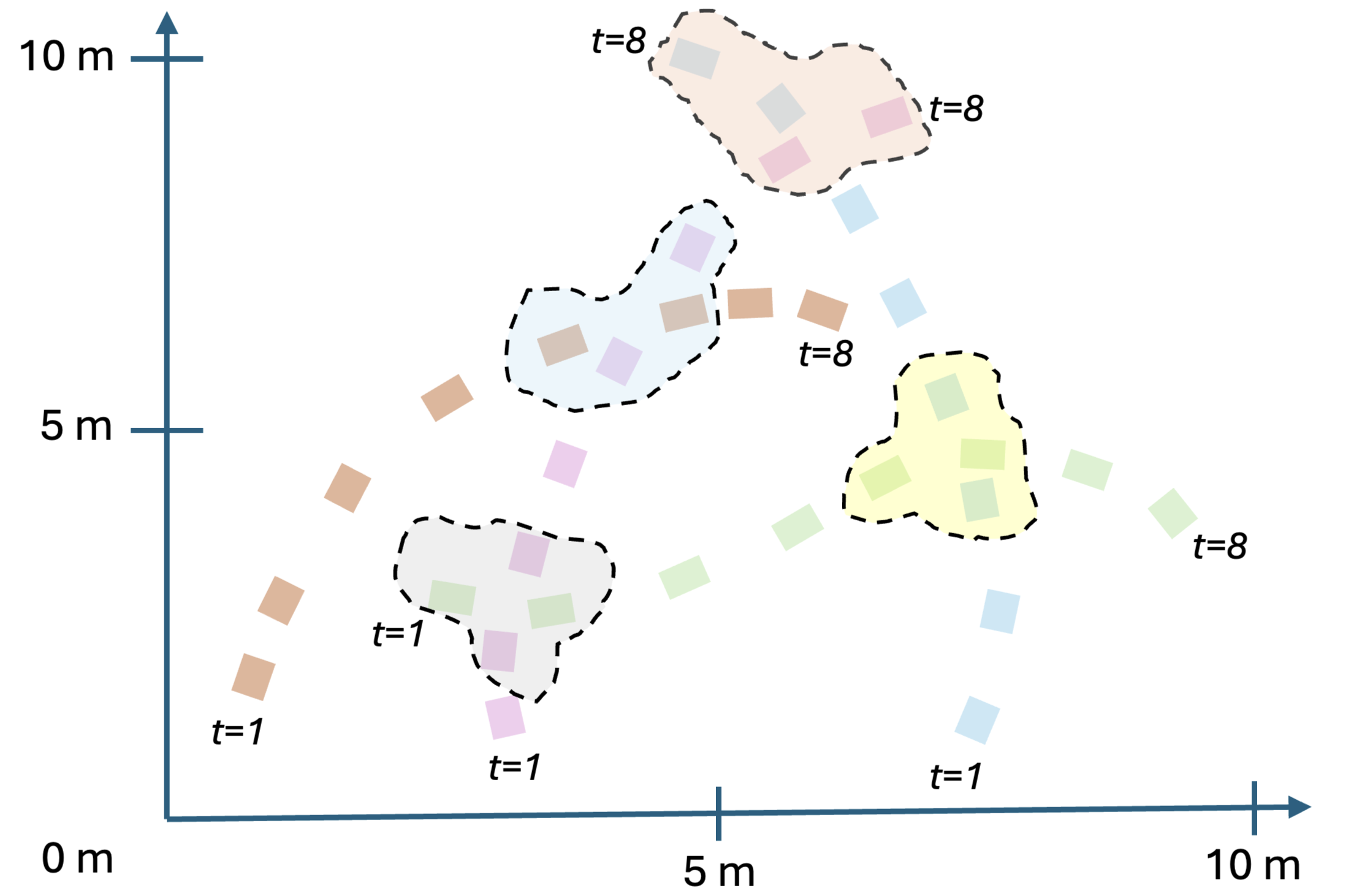}
    \caption{Motion of 4 robots across 8 seconds with interaction zones marked. Within two pose recordings at 1Hz, there can be significant motion and associated interaction of the robots, which are not captured well with discrete-time models, but can be captured more explicitly with continuous time dynamics modeling with PINCoDE. In practice, our model produces much longer goal conditioned forecasts that can span 1-4 minutes encompassing many more interactions between a larger number of agents.}
    \vspace{-3mm}
    \label{fig:robot_motion}
\end{figure}

For a large fleet of robots operating in a shared warehouse environment, communication overheads to a central database can restrict pose recording frequencies to maximum values of 1 Hz or lower. At this frequency, there can be significant movement and robot interactions within a one second time window. Furthermore, pose data is likely to be estimated at different time instants across robots within the time window, fundamentally causing a mismatch when rounded off to the nearest second. With many interacting robots moving in close quarters, these truncation errors can add up over time. To deal with such challenges, Neural Differential Equations (NDEs) \cite{neural_ode,latent_ode} are a promising solution. These neural networks can model the dynamics of the system more explicitly in the form of continuous-time differential equations that learn the smooth temporal evolution of robot trajectories. Additionally, neural ODEs can naturally incorporate physics constraints directly into motion forecasts and handle complex situations like the one shown in Figure \ref{fig:robot_motion}. 

Physics-Informed Neural Networks (PINNs) \cite{pinn_orig} work well for modeling systems where the dynamics evolve in continuous-time based on physical laws that are difficult to precisely model based on known equations. Multi-agent motion forecasting for autonomous robots is one such application where continuous-time NDEs and PINNs can potentially benefit the modeling of robot dynamics. In this work, we explore a variant of NDEs called Neural Controlled Differential Equations (NCDEs) \cite{neural_cde} that we extend to explicitly model how control dynamics laws guide the evolution of robot states. Furthermore, we adopt a curriculum learning strategy to progressively train our continuous-time model over longer horizons while maintaining stability.  

Our primary contributions can be summarized as follows:
\begin{itemize}
    \item  We develop a Physics-Informed Neural Controlled Differential Equations approach (PINCoDE) that is capable of accurate motion forecasting over 1-4 minute horizons for multiple robots, conditioned on goals.
    \item  We introduce physics-informed constraints into the network training and show that it results in significantly improved performance over discrete-time baselines.
    \item  A simple but effective strategy for scaling our method is devised to adapt from our primary experimental space of 10 robots to a larger space of 100 robots operating in the same environment.
    \item A curriculum learning strategy that progressively trains on longer horizons with the PINCoDE model results in a $2.7\times$ improvement over analytical models for motion prediction across a 4 minute horizon.
\end{itemize}

\section{Related Work}
\textbf{Motion Forecasting:} Well established motion forecasting methods such as Trajectron++ \cite{trajectron++} and PreCOG \cite{precog} typically handle prediction of a future time window of a few seconds, with their rollouts for longer time horizons diverging from true states. Neural CDEs \cite{neural_cde} have been explored as a continuous time equivalent of discrete-time RNN models \cite{rnn}. However, the basic neural CDE architecture expects the entire time-series of states to be available at the time of inference. This does not work for robot forecasting tasks where predictions have to be made online in a causal manner for future time steps. An online variant of neural CDE has been introduced in \cite{online_neural_cde}. However, the ``control path" used by the model is the history of the variable being predicted, but not actual controls that guide the state evolution. In this work, we differ from existing approaches by incorporating more explicit controls in terms of reference linear and angular velocities that the robots aim to reach over a long time horizon. 

A majority of motion forecasting literature has been developed for autonomous driving and robotics domains where the intended behavior of external agents such as cars, pedestrians, cyclists, etc., are not known to the ego agent. The use case of a fleet of autonomous mobile robots in a shared environment allows the possibility of using information about partially-known motion plans of all robots to better predict their exact motion. While we can control how we want the robots to move, simulating the movement of hundreds of robots incorporating all aspects of the software stack, and obtaining forecasts can be prohibitively resource intensive and impractical for real-time usage. Hence, motion forecasting taking future reference goals into account is still an important aspect of these systems. \newline

\textbf{Robot Dynamics Simulation:} In a different context, multi-agent reinforcement learning (RL)-based planners and other neural planners rely on data and scenario generation from simulators. These simulators can be slow due to explicit modeling of physics with graphics engines and are expensive to run. The slow speed of such simulators makes it difficult to mine a variety of challenging scenarios. An alternative approach to this problem is the use of a deep learning-based surrogate simulator that learns from real-world data, and can evolve in continuous time. We show that a neural CDE-based trajectory forecasting model which is capable of learning the underlying differential equations can effectively simulate dynamics over long horizons to replicate the capabilities of a much slower simulator. This can be used as a good resource for training RL policies, potentially helping with the sample efficiency problem for rare scenarios. In robotics and RL, the Sim2Real gap \cite{sim2real} describes how models and algorithms fail in the real world for robotic tasks after performing well in simulation. With a well-trained, physics-informed, neural CDE-based surrogate simulator that learns from real-world data, it might also be possible to bridge this gap.  
\newline

\textbf{Physics-informed Machine Learning:}
Physics-informed neural networks \cite{pinn_orig,pinn,chemkin_no,fno_fluid_dynamics} and neural operator learning \cite{neural_operator,fourier_neural_operator,deeponet} are being increasingly studied for their applications in scientific discovery. Neural operators constitute a type of models that learn mappings between infinite-dimensional function spaces. Physics-informed machine learning incorporates physics domain-knowledge based constraints into the learning of machine learning models, allowing them to converge faster and show improved performance. A combination of the two methods has produced ideas such as Physics-Informed Neural Operators (PINOs) \cite{pino,var_pino}, which have helped achieve the best of both worlds. These methods have worked well in various applications in science such as chemical kinetics \cite{phychemnode} and fluid dynamics \cite{fno_fluid_dynamics}. Physics-informed machine learning has been used to solve inverse problems in science that involves identifying causal factors that produce a set of observations \cite{piml_inverse_design,piml_inverse_design_2,piml_inverse_design_3}. It has also been used to build neural surrogate simulators \cite{piml_inverse_design_3, neural_simulator_1} for faster and cheaper high-fidelity simulations. \newline

\textbf{Neural Differential Equations:} 
Neural Differential Equations (NDEs) \cite{neural_ode} are a special case of neural operators in the sense that they aim to learn the explicit differential operator that maps between two function spaces. A common use case of this is to learn a neural differential equation for mapping the input of a dynamical system to its output in continuous time by learning from data. These have been used in dynamics modeling, time-series analysis \cite{latent_ode,neural_cde} and generative models such as normalizing flows \cite{neural_ode}. They have several attractive properties such as being able to directly trade off numerical precision for speed, an adjoint strategy for lowering memory usage and a vector-Jacobian strategy for differentiability through ODE solvers \cite{neural_ode}. Variants of NDEs have been proposed such as variational RNN-ODEs \cite{latent_ode} which learn a distribution that can be sampled from over time, Neural Controlled Differential Equations (NCDEs) \cite{neural_cde} for irregularly timed data, stochastic NDEs \cite{stochastic_de} which use Brownian noise to produce multi-modal predictions and neural rough differential equations \cite{rough_de} that use summaries of local intervals for long time series. \newline
In this work, we develop a variant of neural Controlled Differential Equations (CDEs) that are conditioned on reference controls and are constrained by physics-based domain knowledge for multi-agent systems. We posit that a model which is aware of the underlying physical laws, enforces soft constraints to regulate higher-order derivatives of motion, and learns deviations from those laws using data with a continuous-time differential equation, can stabilize long-horizon forecasts starting from only an initial state.

\section{Physics-informed Goal-Conditioned Neural CDEs for Motion Forecasting}\label{sec:method}
\subsection{Multi-agent System Dynamics}
We aim to model the dynamics of $N$ robots operating in a shared environment. 
The state of each robot $i$ at time $t$ can be modeled by the Special Euclidean group $\mathrm{SE}(2)$, with state
\[
s_{i,t} = (x_{i,t}, y_{i,t}, \theta_{i,t}) \in \mathrm{SE}(2)
\]
The joint state of the multi-agent system composed of all $N$ robots at time $t$ can then be defined as
\[
S_t = (s_{1,t}, \ldots, s_{N,t})
\]
Additionally, to model the dynamics of this system conditioned on goals, we aim to use goal control references over the time horizon as a time-series of goal state derivatives. These control references, which are the reference linear and angular velocities provided to the robots at time $t$, are represented as \[c_t = \{(v_{i,t}, \omega_{i,t})\}_{i \in \{1, 2, ..., N\}}\]Our goal is to learn a differential equation of the form
\[\frac{dS_t}{dt} \;=\; h_\theta \big(S_t, {c_t}\big)\,\]
where $h_\theta$ is a neural network parameterized by $\theta$. 
A sequence of goal reference velocities can typically be obtained in a controllable multi-agent system based on Model Predictive Control (MPC) generated control action references (and target trajectory) for smaller time horizons. For longer time horizons, future reference velocities for time $t$ can be approximated based on reference waypoints of the form \[\{(x_{t_k},\, y_{t_k},\, \theta_{t_k})\}_{k=1}^H,
\quad t < t_1 < t_2 < \cdots < t_{H}\]Based on these waypoints and intended/predicted times of reaching such waypoints, we can estimate approximate velocities that can be used as control signals. 



\subsection{Background for Neural CDEs}
Neural Ordinary Differential Equations (NODEs) are a class of models that are based on solving the Riemann Integral, which can be represented as \[z_t = z_0 + \int_0^t f(z_s) ds\]Here, $f(z_s)$, which represents the time-derivative of the state of the system, is modeled using a neural network and this family of models is compatible with most modern architectures as the choice for $f(z_s)$. The NODE representation can be thought of as a continuous time-equivalent of a Residual Network (ResNet), which is typically represented as \[z_{t+1} = z_t + F(z_t, W_t)\]where $W_t$ is the learnable weight matrix. Neural Controlled Differential Equations (CDEs) are an extension of the neural ODE architecture based on the Riemann-Stieltjes Integral, defined as
\[z_t = z_0 + \int_0^t g(z_s) dX_s\] where $X_s$ is a control input guiding the evolution of the differential equation over time in addition to the derivative of the state with respect to the control input, which is learned. The term $g(z_s)$ here represents the derivative of the state with respect to the control $X_s$ instead of time in the case of the NODE. 
Neural CDEs can be thought of as a continuous-time variant of the RNN. Typically, the control input $X_s$ is a history of past states for the neural CDE, but for our case, we instead use a set of future goal velocities.

\subsection{Model Architecture}
Our PINCoDE model is composed of two components - an autoencoder for learning a joint latent representation for the multi-agent system, followed by a neural CDE that propagates the latent state across time conditioned on future goal velocities. We discuss these in the next subsections, followed by details of how physics constraints are incorporated into the network.
\begin{figure*}[t]
    \centering
    \includegraphics[width=\linewidth]{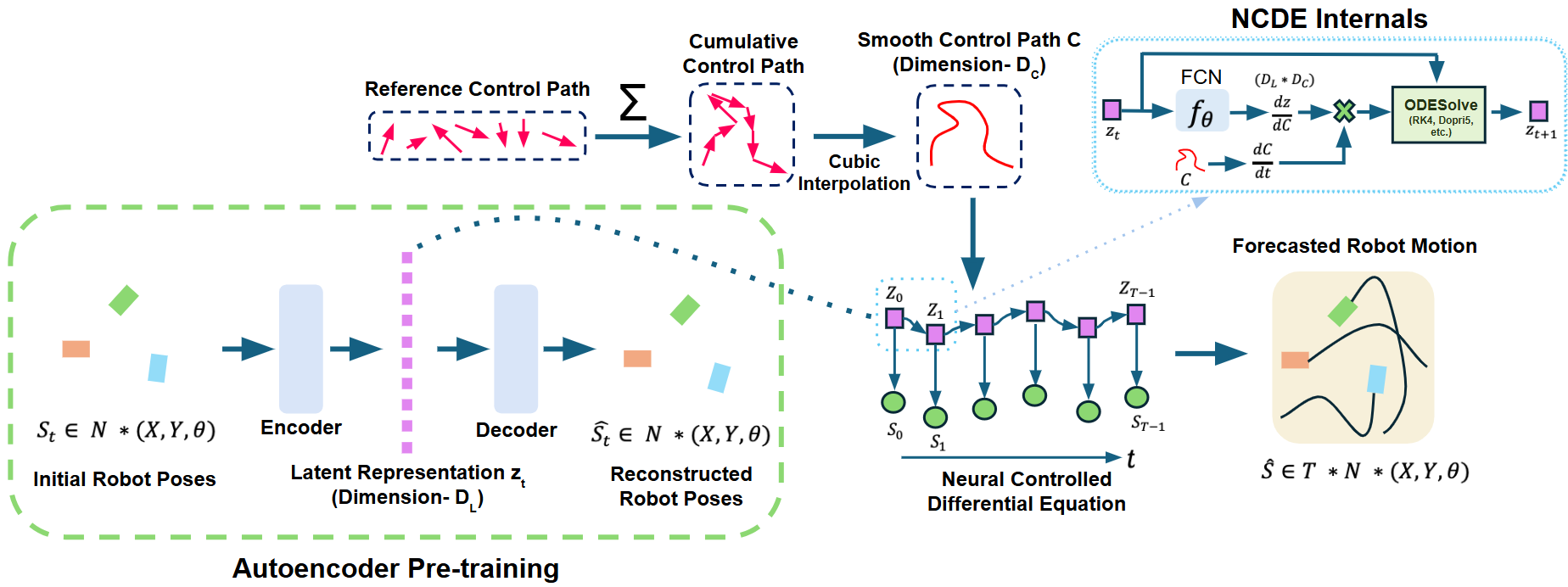}
    \caption{Architecture of our model that uses an autoencoder followed by a latent neural CDE which is additionally guided by reference controls for motion forecasting. $S_t$ and $Z_t$ represent the multi-agent state and latent state at time $t$, $D_L$ refers to the dimension of the learned latent representation and $D_C$ refers to the dimension of the reference controls. FCN refers to a Fully Connected Network and $C$ represents a smoothened version of the control path obtained after taking a cumulative sum of raw reference controls $c_t$.} 
    \label{fig:ncde_plus_ae}
\end{figure*}
\subsubsection{Autoencoder} 

We first train an autoencoder (AE) that takes the poses of multiple robots and learns a rich latent representation which is capable of denoising the input and allows easier training for the subsequent stages of the network. The poses of all the drives are reconstructed from the latent state and this network is trained first, as shown in Figure \ref{fig:ncde_plus_ae}. The encoder and decoder consist of simple 5-layer fully connected networks (FCNs) with hidden layers of size 512 to learn a 30-dimensional latent space for the neural CDE. While this is a 1:1 map in terms of the size of the latent state, we empirically find that the denoised latent state helps the neural CDE learn significantly better. This network is first trained until we obtain high quality reconstructions with robot pose errors of below 0.1m. We observe that a precise reconstruction is critical for the convergence of the subsequent latent CDE.  

\subsubsection{Latent CDE with Control Goal Conditioning}

Once the autoencoder is pretrained, we use the obtained latent representation for training the neural CDE. A sequence of target future velocities is available to the robots at the beginning of the prediction horizon, typically through a combination of direct MPC-generated references for smaller windows (e.g. 5 seconds) and can be derived approximately for longer horizons (1-4 minutes) based on motion goal locations and intermediate waypoints. In practice, for training our model on warehouse robot data recorded over months, we use sparsely recorded actual future linear and angular velocities over the forecasting horizon, which serve as an approximate proxy for reference goal velocities. During inference and deployment, MPC generated goals can be used for shorter horizons and reference velocities based on waypoints for the longer ones, potentially with more finetuning. Additionally, since CDEs work on continuous control paths, we take a cumulative sum of these reference controls over time, fit a piecewise Hermite cubic spline with backward differences to generate a smooth and differentiable curve, and then use this reference to predict the movement of robots over time. 


The neural network $f_\theta$ from the NCDE internals block in Figure \ref{fig:ncde_plus_ae}, is a fully connected Multi-layer Perceptron (MLP) with 4 hidden layers of width 512, and is learned from data. The parameters of this network represent the derivative of the underlying differential equation that models how the state changes with respect to the control inputs. This derivative  $\frac{dz}{dC}$ from Figure \ref{fig:ncde_plus_ae} is combined with the time-derivative of the smooth reference control path $\frac{dC}{dt}$ with chain rule of differentiation. The neural differential equation is then used as part of a standard ODE solver such as the fourth-order Range-Kutta for integration over time to predict the next latent state for the next second. Internally, the forward pass for the network is run several times at the evaluation points of the ODE solver to produce an accurate approximation of the continuous-time integral. These states are then passed through the pre-trained decoder to reconstruct individual robot poses from the joint state.

\subsection{Physics-Informed Loss} 

We first use a Mean Squared Error (MSE) reconstruction loss for the AE, followed by a weighted combination of an MSE future pose prediction loss, a physics-informed unicycle loss for dynamic feasibility and an acceleration regularization loss to ensure smooth transitions in velocity across time steps. The pose prediction loss is defined as \[L_{pred} = \|S_{1:T-1} - \hat{S}_{1:T-1}\|^2\], where $S$ and $\hat{S}$ represent the ground truth and predicted states for the multi-agent system across the time horizon $T$. Unicycle dynamics are represented by the simple model governed by 
\[
\begin{aligned}
    \dot{x} = v \cos \theta,
    \dot{y}=v \sin \theta, 
     \dot{\theta} = \omega
\end{aligned}
\]
where:
\begin{itemize}
    \item \(x, y\) represent the 2D location of the robot
    \item \(\theta\) represents the heading angle of the robot
    \item \(v\) is the linear velocity of the robot
    \item \(\omega\) is the angular velocity of the robot
\end{itemize}
\begin{figure*}[t]
    \centering
    \includegraphics[width=0.85\linewidth]{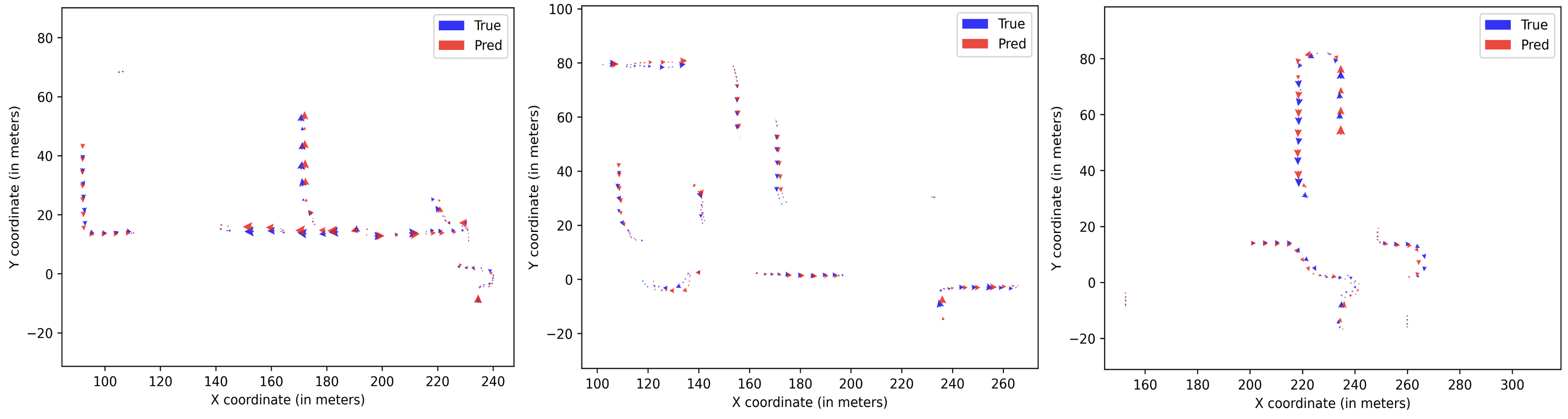}
    \label{fig:pred1}
    \vspace{-2mm}
    \caption{Three instances of motion forecasting performance over a 60 second time horizon. Blue arrows show ground truth and red arrows show prediction, with strong correspondences observed over long time horizons.}
    \label{fig:predictions}
\end{figure*}
Based on this, the unicycle loss can be represented by the following equation. \[L_{uni} = \sum_{t=0}^{T-2} [(\dot{x}_t - v_t \cos \theta_t)^2 + (\dot{y}_t - v_t \sin \theta_t)^2 + (\dot{\theta}_t - \omega_t)^2]\] enforcing adherence to the physical constraints between the independently predicted velocity and angle values. Acceleration regularization is captured by \[
L_{acc}=\frac{1}{T-1} \sum_{t=0}^{T-2} \left( \frac{\lVert\hat v_{t+1}\rVert - \lVert \hat v_t \rVert}{\Delta t} \right)^{2}\], to ensure there are no sudden jumps between consecutive predictions and the learned motion is representative of true state. In summary, for the physics-informed training of the PINCoDE model, we use the combined Physics-Informed Forecasting Loss 
\[
\begin{split}
L_{PIFL} &= W_{pred} L_{pred} + W_{uni} L_{uni} + W_{acc} L_{acc} \\
& \quad \quad W_{pred}, W_{uni}, W_{acc} > 0
\end{split}
\]
\section{Experiments}
Our experimental setting is described in this section in terms of the data used, the training strategy, and  specific implementation details.
\subsection{Dataset} 
We use one month of data from a warehouse floor application comprised of multiple robots moving in an unstructured environment. This dataset records poses and other relevant information for many robots every second. For this study, we divide the data into four minute chunks with a slide length of 10 seconds and consider the top N robots with the most movement within this time window. This strategy is adopted to ensure that there is significant robot movement in our dataset, along with overlaps that effectively work as data augmentation. For most of our experiments, we choose $N=10$, considering the top $10$ robots which move the most, except for the scalability analysis experiments where we use the top $100$ robots from the shared environment. 

For the actual warehouse robots, MPC generated velocity references are not recorded in historical data, and hence, we make an approximation to use the future linear and angular velocities $(v,\omega)$ instead. During deployment on an actual robot, the future MPC references can be used along with goals/waypoints as discussed in Section \ref{sec:method}.

Pose data from various robots are estimated and recorded at different times within each second. Since such non-uniformity can hinder model learning capability, we fit cubic splines for each dimension of the robot data- $(x,y,\theta,v,\omega)$ to represent continuous variations. The cubic splines for each pose dimension across all robots are then sampled at the top of every second to create a more uniform dataset at 1Hz. For actual input to our models, these are re-interpolated to create a continuous path. We have a total of about 250,000 sequences with each sequence consisting of 60 seconds of poses, linear and angular velocities for a large number of robots, from which the top 10 most moving ones are used for all of our experiments except for the scalability tests, which use the top 100 robots that move the most. 

\subsection{Training Settings} 

We perform an 80:20 train-to-val split of the data for all of our experiments. Effectively, 24 days of robot data are used for training and the remaining 6 days are used for validation. Such a separation ensures that validation is performed on unseen data with no mixing. We use the commonly used Average Displacement Error (ADE) metric for evaluation. The metric can be defined as follows: \[
\mathrm{ADE} \;=\; \frac{1}{N T} \sum_{i=1}^{N} \sum_{t=0}^{T-1} 
\sqrt{ \big( \hat{x_{i,t}} - x_{i,t} \big)^{2} \;+\; \big( \hat{y_{i,t}} - y_{i,t} \big)^{2} }
\] 
where, $N$ is the number of robots, $T$ is the length of the time horizon, $(\hat{x},\hat{y})$ are the predicted coordinates, and $(x,y)$ represent the ground truth.

\subsection{Implementation Details} 
\textbf{Neural CDE:} We use the TorchDiffEq \cite{neural_ode} and TorchCDE \cite{neural_cde} frameworks for implementing our PINCoDE model. The adjoint method, which is often used in ODEs \cite{neural_ode} to reduce memory usage, is disabled in our case since it leads to slower training. Furthermore, the fourth-order Range-Kutta (RK4) integrator \cite{dormand_prince} is used as the ODE solver with the CDE. Our experiments show that using more modern adaptive solvers such as Dormand-Prince (DOPRI5) \cite{dormand_prince} results in underflow for calculating the time step in stiff regions that form occasionally. Stiff solvers such as LSODA \cite{lsoda} can help in such regimes but lack GPU support in TorchDiffeq library \cite{neural_ode}. In the future, faster implementations in JAX such as Diffrax \cite{kidger_thesis} can be explored for further speed-up during training.

\textbf{Model Training:} We have implemented PINCoDE in Pytorch with DDP parallelism across multiple GPUs. For training, we use the ADAM optimizer \cite{adam_optimizer} with an initial LR of $3\times10^{-4}$ with a Cosine Annealing Scheduler \cite{cosine_annealing}, eventually decaying to $3\times10^{-7}$ over 800 epochs. A batch size of 1600 is used for our experiments. Training our model on the full dataset takes about 2 days on 8 Nvidia A10G GPUs. For training with the physics-informed loss, we empirically choose $W_{uni}=10$, $W_{acc}=10$ and $W_{pred}=1$, which provides a good balance for the training regime where the pose errors reach the $1-2\,m$ range for a 60 second window, and learning complexity increases.
\section{Results and Analysis}
In this section, we present the results of various experiments conducted that show the effectiveness of the PINCoDE approach. These are detailed in the following sub-sections.

\subsection{Visual analysis of predicted trajectories}
Three instances of predictions from our neural CDE-based forecasting model in 10-robot scenarios of varying complexity are shown in Figure \ref{fig:predictions}, where arrows indicate the direction of travel. The red arrows show reconstructed validation poses, while the blue ones are the ground truth. It can be seen that the predicted poses match very well with the ground truth, with accurate directions of the arrows. Furthermore, the size of each arrow is proportional to the velocity magnitude, showing strong correspondence between the dynamics of the predictions and ground truth. Although poses are predicted at 1Hz and can be sampled at any intermediate time, the arrows are drawn once every 5 seconds for clarity.
\begin{table}
  \centering
    \begin{tabularx}{0.5\textwidth}{c*{4}{>{\centering\arraybackslash}X}}
        \toprule
        \textbf{Method} & \textbf{ADE (5s)} & \textbf{ADE (10s)} & \textbf{ADE (60s)} & \textbf{\#Params}\\
        \midrule
        GRU Baseline \cite{gru_paper} & 0.29 m & 0.39 m & 1.22 m & 1.2M \\
        LSTM Baseline \cite{lstm_paper} & 0.25 m & 0.30 m & 0.88 m & 1.5M \\
        TCN Baseline \cite{tcn_paper} & 0.22 m & 0.27 m & 0.84 m & 1.5M \\
        Transformer Baseline \cite{transformer_paper}  & - & - & 1.33 m & 1.4M \\
        \midrule
        \textbf{PINCoDE - Ours} & \textbf{0.18\,m} &\textbf{ 0.24 m} & \textbf{0.77\,m} &  1.1M$^*$ \\
        \bottomrule
    \end{tabularx}
  \caption{Comparison of PINCoDE performance against discrete-time motion forecasting baselines. $^*$ Number of active parameters and 2.1M Frozen Parameters for the AE}
  \vspace{-5mm}
  \label{tab:baseline_comparison}
\end{table}
\begin{figure*}[t]
    \centering
    \includegraphics[width=\linewidth]{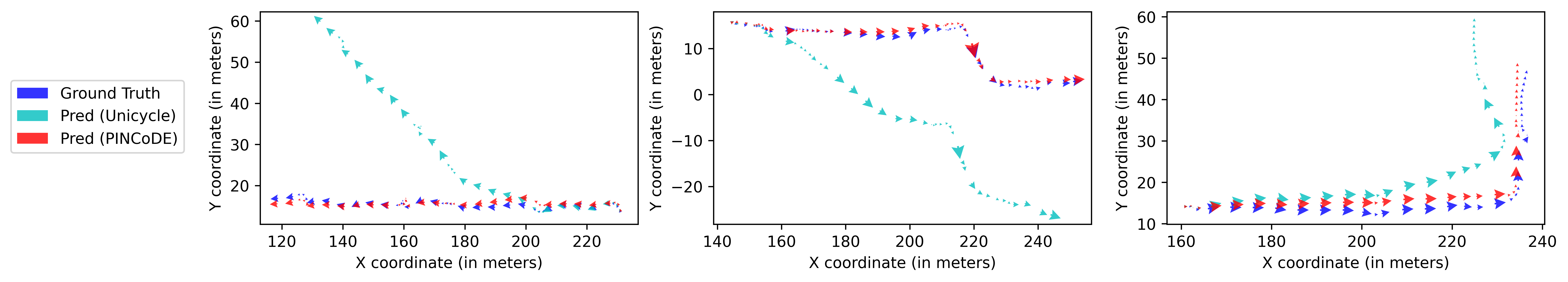}
    \caption{A few instances of control-conditioned motion forecasting where the unicycle model (cyan) heavily diverges compared to the ground truth (blue) while predictions from our PINCoDE model (red) trained with curriculum learning remain close to the ground truth in a 4 minute horizon.}
    \label{fig:pincde_vs_unicycle_traj}    
\end{figure*}
\subsection{Performance comparison}
For a structured evaluation of model performance, we compare our model against other baseline models that fit well for our trajectory forecasting task. It should be noted that many of the well-studied trajectory forecasting methods use history from the last few seconds to make predictions for the next 5-10 seconds. Our problem setting is different from this since we only use the initial states and future references, without history, to make it suitable for use even in surrogate simulation. Furthermore, we use goal conditioning based on reference velocities to be tracked for making forecasts over much longer horizons. This is not the case with a lot of existing literature where the behavior of each agent is not as controllable as in our case. Hence, we compare our method with suitable strong baselines that are capable of modeling the time series of reference controls and can incorporate these into making predictions at each step throughout the time horizon starting from the same initial condition.

The AE of our PINCoDE model is first trained to obtain a pose reconstruction error of $7 cm$ on average. This is followed by freezing the AE and only training the $1.1$ million parameter neural CDE. Table \ref{tab:baseline_comparison} shows the performance of our model compared to strong baselines that use discrete-time models. We use a Gated Recurrent Unit (GRU) \cite{gru_paper}, a Long Short Term Memory (LSTM) network \cite{lstm_paper} and a Temporal Convolution Network (TCN) \cite{tcn_paper} as our baselines. These model the time series of reference controls at each time step, and evolve the joint state of all robots in discrete time starting from the initial poses, similar to our proposed neural CDE model. Furthermore, we also use a transformer model \cite{transformer_paper} that can attend to the entire sequence of control inputs at once and then auto-regressively decode to propagate the multi-agent state in discrete time. It is to be noted that the transformer has a fixed sequence length and cannot be directly evaluated on smaller windows. The parameter counts for the baseline models are chosen to keep them comparable to the number of active parameters in our PINCoDE model. All the baseline models are trained directly to predict the residuals at each time step starting from the initial poses. We observe that the PINCoDE model obtains a forecasting Average Displacement Error (ADE) of $0.77 m$ in a $60$ second time horizon, performing better than the discrete time baselines. Among the baselines, the TCN network works the best, obtaining an ADE of $0.84m$. Similar trends are observed when the same models are used to infer in shorter horizons of $5$ seconds and $10$ seconds, without further training. 
\begin{table}
    \centering
        \begin{tabular}{c|c|ccc}
        \toprule
        Goal Controls & Physics & \textbf{ADE (5s)} & \textbf{ADE (10s)} & \textbf{ADE (60s)}\\
        \midrule
        $\times$ & $\times$ & 0.59 m & 1.28 m & 7.31\,m\\
        $\times$ & $\checkmark$ & 0.55 m & 1.22 m & 7.29\,m \\
        \midrule
        $\checkmark$ & $\times$ & 0.21\,m & 0.30 m & 0.93\,m\\
        $\checkmark$ & $\checkmark$ & \textbf{0.18\,m} & \textbf{0.24 m} &   
        \textbf{0.77\,m}\\
        \bottomrule
        \end{tabular}
        \caption{Ablation study for the PINCoDE model.}
        \vspace{-5mm}
        \label{tab:ablation}
\end{table}
\subsection{Ablation studies}

Table \ref{tab:ablation} compares the impact of control references as goals, and the effect of incorporating physics constraints into the training of our PINCoDE model. It should be noted that the neural CDE model without control conditioning becomes equivalent to a neural ODE model \cite{neural_ode}. It can be concluded from Table \ref{tab:ablation} that without the reference controls incorporated as part of the model, it is difficult to predict how the robots will move since there are many possible movement configurations for the same initial state. With goal controls, performance improves significantly. PINCoDE is especially useful in environments where the target locations of the robot are known beforehand, such as fleets of mobile robots in warehouses. Controls are used in the discrete-time models as well for our analysis, but the impact of continuous-time modeling can also be seen with our PINCoDE model as compared to discrete-time equivalents in Table \ref{tab:baseline_comparison}. Additionally, based on Table \ref{tab:ablation}, we see that physics-informed losses help explicitly enforce physical constraints that guide robot motion, and this leads to enhanced generalization.
\begin{figure}
    \centering
    \includegraphics[width=0.75\linewidth]{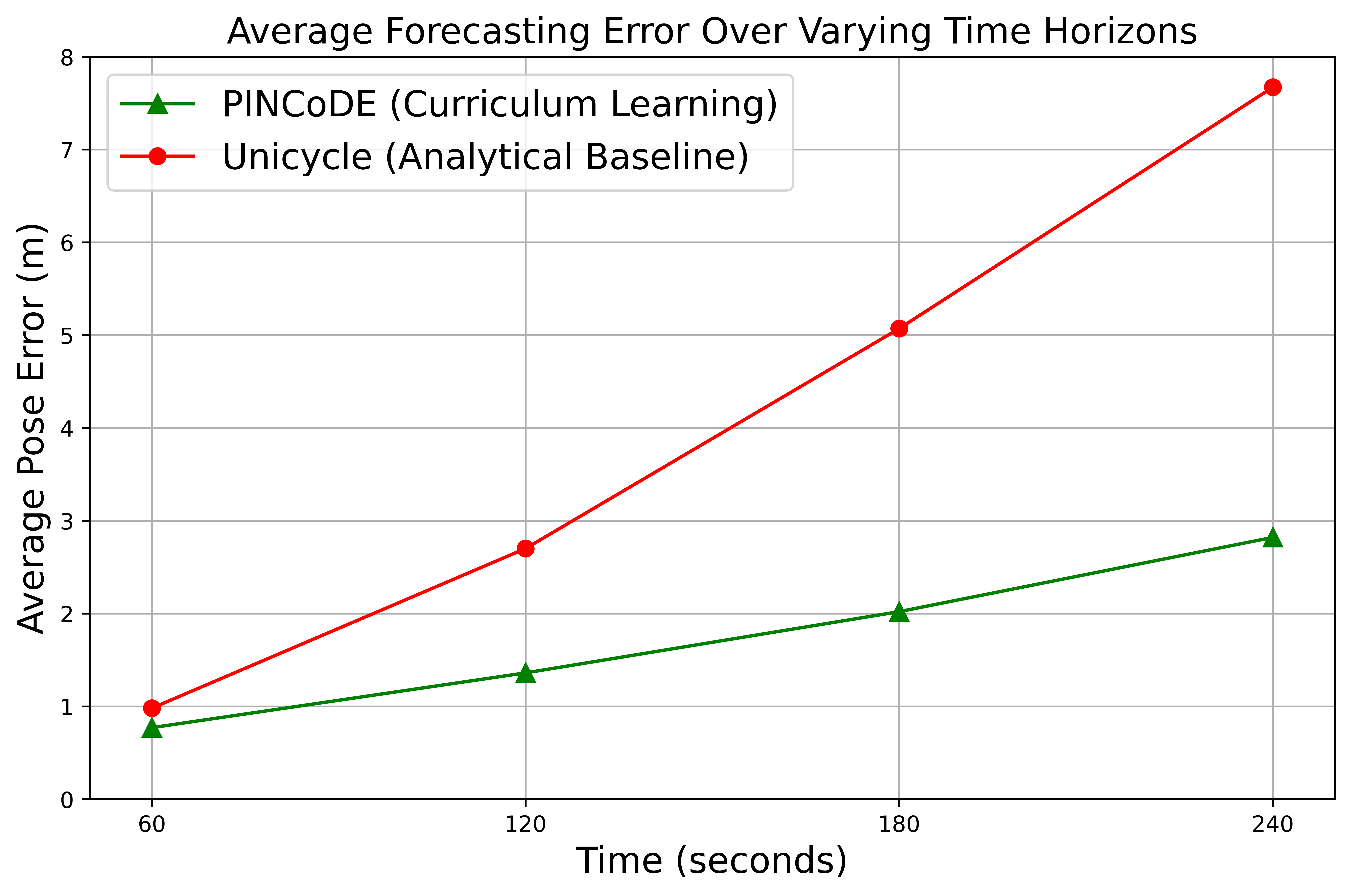}
    \caption{PINCoDE with curriculum learning for progressively training across longer horizons results in a significantly lower error as compared to analytical models like the unicycle.}
    \vspace{-5mm}
    \label{fig:curriculum_learning}
\end{figure}
\subsection{Curriculum Learning}
To extend our model to produce non-divergent forecasts for time horizons longer than 60 seconds, like 4 minutes, we adopt a curriculum learning strategy \cite{curriculum_learning}. Such long time horizons can be useful for surrogate simulators where the goal is to produce historically-consistent rollouts for multiple agents from initial states to generate data or study algorithmic impact. For curriculum learning, we start with our best performing physics-informed model from the 60 second horizon and fine-tune it sequentially for 120 seconds, 180 seconds and 240 seconds. Figure \ref{fig:curriculum_learning} shows the relative performance gains against an analytical unicycle model baseline that would typically be used in dynamical systems to forward propagate the robot state based on velocity references. The difference in performance is already noticeable at the 60 second time horizon in Figure \ref{fig:curriculum_learning} with a value of $0.77\,m$ with PINCoDE compared to a $0.98\,m$ error of the unicycle model, showing a $27.3\%$ gain. However, for the 4-minute forecasting window, our PINCoDE model with curriculum learning obtains an average prediction error of $2.82\,m$ compared to $7.67\,m$ from the unicycle model, showing a much larger $2.7\times$ improvement. Learning the dynamics of motion from data, aided by physics-based constraints, helps stabilize predictions significantly. 

Figure \ref{fig:pincde_vs_unicycle_traj} shows this effect visually for a few examples where the unicycle model prediction over a 4 minute window diverges heavily as time progresses, but PINCoDE predictions stay close to the ground truth. The robot yaw can change quite fast when turning and this does not get captured well with analytical unicycle models. However, PINCoDE is able to specifically learn these nuances from the data and adjust the differential equation appropriately to deal with such situations.

\begin{figure}
  \centering
  \includegraphics[width=0.8\linewidth]{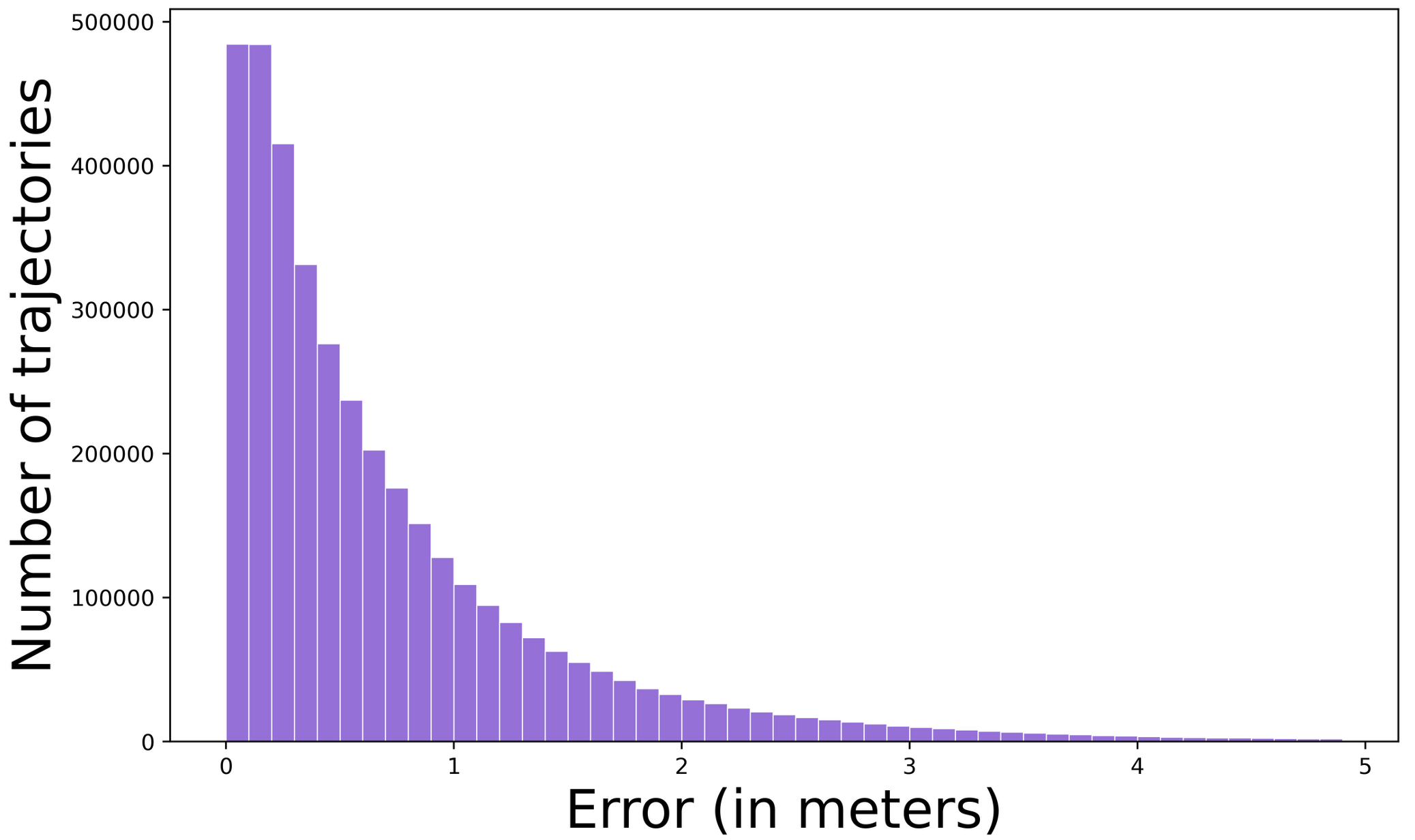}
  \caption{Distribution of errors from our neural CDE model}
  \label{fig:errors_histogram}
\end{figure}


\subsection{Distribution of Errors}
In Figure \ref{fig:errors_histogram}, we plot the distribution of error predictions for each robot trajectory. While the best ADE from Table 1 has a value of 0.77 m, the actual distribution of forecasting errors shows that for a large number of cases, the prediction errors are much lower, less than 0.4 m. Moreover, there are very few instances with large errors. 
\subsection{Runtime}
We also quantify the runtime of our PINCoDE model on a single Nvidia A10G GPU with 24GB memory. A batch of 2048 sequences, each with 10 robots and 60 seconds, takes less than 1 second for inference, paving the way for a parallelizable, fast and differentiable surrogate simulator.

\subsection{Scalability to Large-scale Problems}
The final set of experiments aims to study the scalability of our approach to much larger multi-agent systems. Specifically, we now consider a system of 100 robots, expanding upon earlier training with 10 robots. Among the options for scaling, the first is to train the PINCoDE model from scratch with the same architecture while increasing the input dimensions. Our experiments show that reconstruction accuracy suffers considerably for such a large input space, even on increasing the model capacity, and it also takes much longer to train. Instead, motivated by the fact that our model for 10 robots runs much faster than the speed needed for real-time operation, for the 100 robot use case, we break it down into 10 groups of 10. The model is then run on these 10 subsets of robot and their predictions are combined. To best capture the interactions between the robots and use them to influence model training, we cluster the 100 robots based on their initial positions into groups of 10, and combine the predictions produced by running the previously trained model on each group.
\begin{table}
        \centering  
        \begin{tabular}{cccc}
        \toprule
        \textbf{Method} & \textbf{ADE (5s)} & \textbf{ADE (10s)} & \textbf{ADE (60s)}\\
        \midrule
        {Zero-shot PINCoDE} & 0.12\,m & 0.16\,m & 0.54\,m \\
        {Fine-tuned PINCoDE} & \textbf{0.11\,m} & \textbf{0.15\,m} & \textbf{0.46\,m} \\        
        \bottomrule
        \end{tabular}
        \caption{Scalability test for our PINCoDE model to predict for 100 robots using the model pre-trained on 10 robots.}
        \label{tab:scalability}
\end{table}



\begin{figure}[t]
    \centering
    \includegraphics[width=\linewidth]{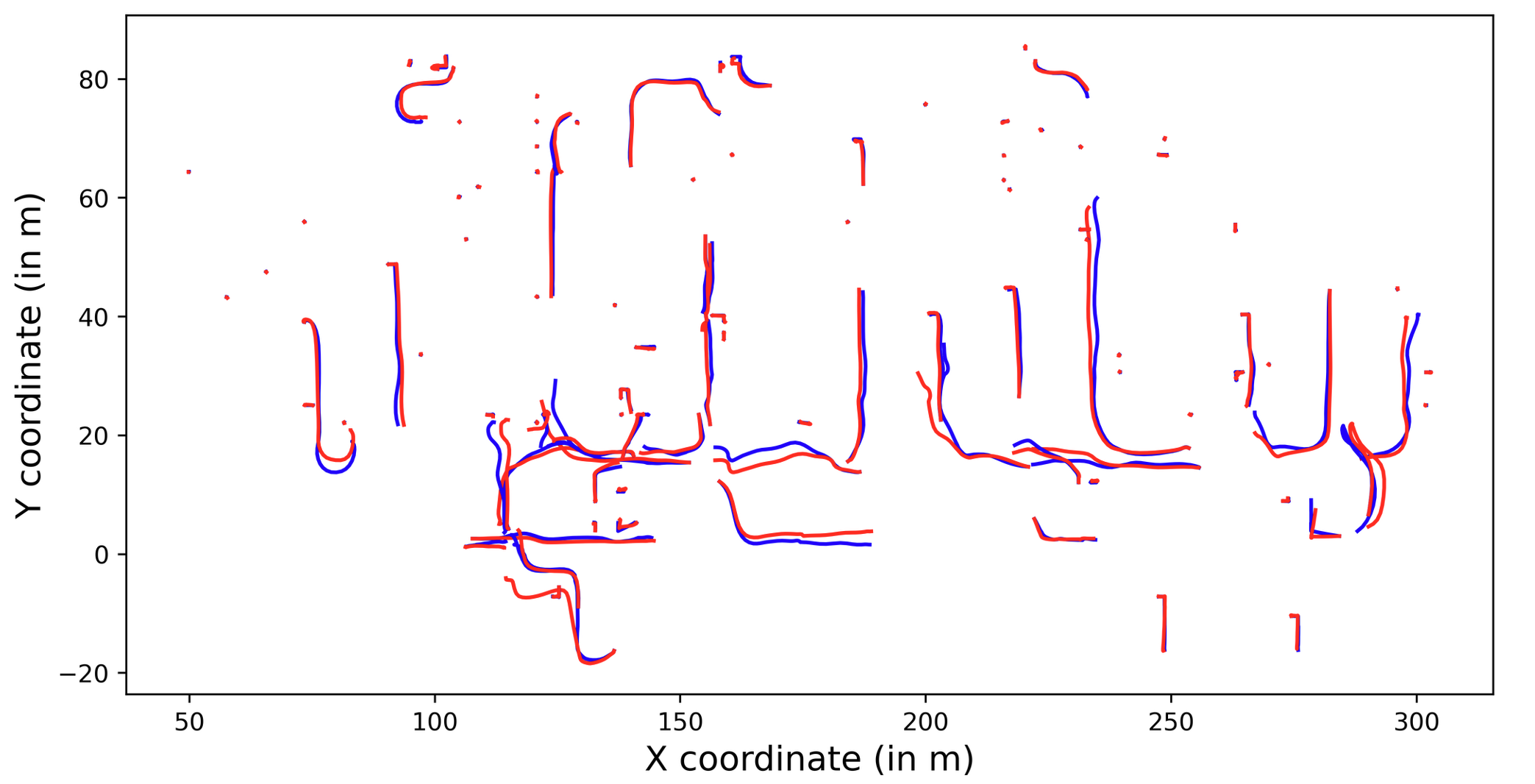}
    \caption{Motion prediction for a 60 second horizon for 100 robots with our scaling approach for PINCoDE.}
    \vspace{-3mm}
    \label{fig:100_robots_prediction}
\end{figure}

Table \ref{tab:scalability} shows how how PINCoDE initially trained on 10 robots performs on the larger scale of 100 robots. Without any further training on the 100 robot data, in a zero-shot setting, our model already obtains a competitive ADE of $0.54m$ in 60 seconds. On further finetuning with weights shared across the 100 robots in 10 robot chunks, the error goes down to $0.46m$. The actual error obtained for 100 robots is lower than the one for 10 robots, which is counter-intuitive. On analyzing the 100 robot data, we make an additional observation that, on average, no more than $50\%$ of the robots are mobile within each one minute time window. This explains the observed lower average error compared to the 10 robot case since it is much easier to predict no movement compared to an actual trajectory over a long window. Figure \ref{fig:100_robots_prediction} shows that the predicted red trajectories match up very well with the blue ground truth in most cases, showing consistent predictions for 100 robots. This establishes the fact that our approach is easy to scale, without using a larger model, and scaling can be quite effective in practice. 

\section{Conclusion}
Motion forecasting over long horizons for multi-agent systems is a challenging but important problem. In this work, we have developed a method named PINCoDE that can model the joint latent dynamics of multiple robots in continuous time over long time horizons of 60 seconds. Our method uses neural controlled differential equations which incorporate goal velocities over time as the control path for the neural CDE. We explore the capabilities of these models for motion forecasting as opposed to more commonly used transformer/RNN-based forecasting models. PINCoDE results show that continuous-time methods with goal conditioning in terms of reference controls are highly effective for motion forecasting over long horizons. Furthermore, adding a physics-informed loss significantly improves the quality of rollouts. Additionally, a simple scaling strategy from 10 robots to 100 robots produces consistent predictions without the need for extra model capacity. We show that using a curriculum learning strategy with our Physics-informed Neural CDE model results in a $2.7\times$ reduction of forecasted pose error over time horizons as long as 4 minutes compared to analytical models. However, a key limitation of this work is that we do not model the effect of uncontrollable agents such as pedestrians in the environment, which can also have an impact on how the robots move. Future extensions can incorporate such agents into our model, borrowing ideas from prior literature on pedestrian behavior prediction.
\bibliographystyle{IEEEtran}
\bibliography{references} 

\begin{thebibliography}{10}
\providecommand{\url}[1]{#1}
\csname url@samestyle\endcsname
\providecommand{\newblock}{\relax}
\providecommand{\bibinfo}[2]{#2}
\providecommand{\BIBentrySTDinterwordspacing}{\spaceskip=0pt\relax}
\providecommand{\BIBentryALTinterwordstretchfactor}{4}
\providecommand{\BIBentryALTinterwordspacing}{\spaceskip=\fontdimen2\font plus
\BIBentryALTinterwordstretchfactor\fontdimen3\font minus \fontdimen4\font\relax}
\providecommand{\BIBforeignlanguage}[2]{{%
\expandafter\ifx\csname l@#1\endcsname\relax
\typeout{** WARNING: IEEEtran.bst: No hyphenation pattern has been}%
\typeout{** loaded for the language `#1'. Using the pattern for}%
\typeout{** the default language instead.}%
\else
\language=\csname l@#1\endcsname
\fi
#2}}
\providecommand{\BIBdecl}{\relax}
\BIBdecl

\bibitem{baniodeh2025scalinglawsmotionforecasting}
\BIBentryALTinterwordspacing
M.~Baniodeh, K.~Goel, S.~Ettinger, C.~Fuertes, A.~Seff, T.~Shen, C.~Gulino, C.~Yang, G.~Jerfel, D.~Choe, R.~Wang, V.~Kallem, S.~Casas, R.~Al-Rfou, B.~Sapp, and D.~Anguelov, ``Scaling laws of motion forecasting and planning -- a technical report,'' 2025. [Online]. Available: \url{https://arxiv.org/abs/2506.08228}
\BIBentrySTDinterwordspacing

\bibitem{wayformer}
N.~Nayakanti, R.~Al-Rfou, A.~Zhou, K.~Goel, K.~S. Refaat, and B.~Sapp, ``Wayformer: Motion forecasting via simple and efficient attention networks,'' in \emph{ICRA}, 2023, pp. 2980--2987.

\bibitem{trajectron++}
T.~Salzmann, B.~Ivanovic, P.~Chakravarty, and M.~Pavone, ``Trajectron++: Dynamically-feasible trajectory forecasting with heterogeneous data,'' in \emph{ECCV}, 2020, p. 683–700.

\bibitem{motionlm}
\BIBentryALTinterwordspacing
A.~Seff, B.~Cera, D.~Chen, M.~Ng, A.~Zhou, N.~Nayakanti, K.~S. Refaat, R.~Al-Rfou, and B.~Sapp, ``Motionlm: Multi-agent motion forecasting as language modeling,'' 2023. [Online]. Available: \url{https://arxiv.org/abs/2309.16534}
\BIBentrySTDinterwordspacing

\bibitem{compounding_error}
N.~Lambert, K.~Pister, and R.~Calandra, ``Investigating compounding prediction errors in learned dynamics models,'' \emph{arXiv preprint arXiv:2203.09637}, 2022.

\bibitem{neural_ode}
R.~T.~Q. Chen, Y.~Rubanova, J.~Bettencourt, and D.~Duvenaud, ``Neural ordinary differential equations,'' in \emph{NeurIPS}, 2018, p. 6572–6583.

\bibitem{latent_ode}
\BIBentryALTinterwordspacing
Y.~Rubanova, R.~T.~Q. Chen, and D.~Duvenaud, ``Latent odes for irregularly-sampled time series,'' 2019. [Online]. Available: \url{https://arxiv.org/abs/1907.03907}
\BIBentrySTDinterwordspacing

\bibitem{pinn_orig}
\BIBentryALTinterwordspacing
G.~E. Karniadakis, I.~G. Kevrekidis, L.~Lu, P.~Perdikaris, S.~Wang, and L.~Yang, ``Physics-informed machine learning,'' \emph{Nature Reviews Physics}, vol.~3, no.~6, pp. 422--440, 2021. [Online]. Available: \url{https://doi.org/10.1038/s42254-021-00314-5}
\BIBentrySTDinterwordspacing

\bibitem{neural_cde}
P.~Kidger, J.~Morrill, J.~Foster, and T.~Lyons, ``Neural controlled differential equations for irregular time series,'' \emph{Advances in neural information processing systems}, vol.~33, pp. 6696--6707, 2020.

\bibitem{precog}
\BIBentryALTinterwordspacing
N.~Rhinehart, R.~McAllister, K.~Kitani, and S.~Levine, ``Precog: Prediction conditioned on goals in visual multi-agent settings,'' 2019. [Online]. Available: \url{https://arxiv.org/abs/1905.01296}
\BIBentrySTDinterwordspacing

\bibitem{rnn}
\BIBentryALTinterwordspacing
A.~Sherstinsky, ``Fundamentals of recurrent neural network (rnn) and long short-term memory (lstm) network,'' \emph{Physica D: Nonlinear Phenomena}, vol. 404, p. 132306, Mar. 2020. [Online]. Available: \url{http://dx.doi.org/10.1016/j.physd.2019.132306}
\BIBentrySTDinterwordspacing

\bibitem{online_neural_cde}
\BIBentryALTinterwordspacing
J.~Morrill, P.~Kidger, L.~Yang, and T.~Lyons, ``Neural controlled differential equations for online prediction tasks,'' 2021. [Online]. Available: \url{https://arxiv.org/abs/2106.11028}
\BIBentrySTDinterwordspacing

\bibitem{sim2real}
W.~Zhao, J.~P. Queralta, and T.~Westerlund, ``Sim-to-real transfer in deep reinforcement learning for robotics: a survey,'' in \emph{2020 IEEE Symposium Series on Computational Intelligence (SSCI)}, 2020, pp. 737--744.

\bibitem{pinn}
\BIBentryALTinterwordspacing
M.~Raissi, P.~Perdikaris, and G.~Karniadakis, ``Physics-informed neural networks: A deep learning framework for solving forward and inverse problems involving nonlinear partial differential equations,'' \emph{Journal of Computational Physics}, vol. 378, pp. 686--707, 2019. [Online]. Available: \url{https://www.sciencedirect.com/science/article/pii/S0021999118307125}
\BIBentrySTDinterwordspacing

\bibitem{chemkin_no}
\BIBentryALTinterwordspacing
S.~Goswami, A.~D. Jagtap, H.~Babaee, B.~T. Susi, and G.~E. Karniadakis, ``Learning stiff chemical kinetics using extended deep neural operators,'' \emph{Computer Methods in Applied Mechanics and Engineering}, vol. 419, p. 116674, 2024. [Online]. Available: \url{https://www.sciencedirect.com/science/article/pii/S0045782523007971}
\BIBentrySTDinterwordspacing

\bibitem{fno_fluid_dynamics}
\BIBentryALTinterwordspacing
W.~Xiao, T.~Gao, K.~Liu, J.~Duan, and M.~Zhao, ``Fourier neural operator based fluid–structure interaction for predicting the vesicle dynamics,'' \emph{Physica D: Nonlinear Phenomena}, vol. 463, p. 134145, 2024. [Online]. Available: \url{https://www.sciencedirect.com/science/article/pii/S0167278924000964}
\BIBentrySTDinterwordspacing

\bibitem{neural_operator}
N.~Kovachki, Z.~Li, B.~Liu, K.~Azizzadenesheli, K.~Bhattacharya, A.~Stuart, and A.~Anandkumar, ``Neural operator: learning maps between function spaces with applications to pdes,'' \emph{J. Mach. Learn. Res.}, vol.~24, no.~1, Jan. 2023.

\bibitem{fourier_neural_operator}
\BIBentryALTinterwordspacing
Z.~Li, N.~Kovachki, K.~Azizzadenesheli, B.~Liu, K.~Bhattacharya, A.~Stuart, and A.~Anandkumar, ``Fourier neural operator for parametric partial differential equations,'' 2021. [Online]. Available: \url{https://arxiv.org/abs/2010.08895}
\BIBentrySTDinterwordspacing

\bibitem{deeponet}
\BIBentryALTinterwordspacing
L.~Lu, P.~Jin, G.~Pang, Z.~Zhang, and G.~E. Karniadakis, ``Learning nonlinear operators via deeponet based on the universal approximation theorem of operators,'' \emph{Nature Machine Intelligence}, vol.~3, no.~3, 03 2021. [Online]. Available: \url{https://www.osti.gov/biblio/2281727}
\BIBentrySTDinterwordspacing

\bibitem{pino}
\BIBentryALTinterwordspacing
Z.~Li, H.~Zheng, N.~Kovachki, D.~Jin, H.~Chen, B.~Liu, K.~Azizzadenesheli, and A.~Anandkumar, ``Physics-informed neural operator for learning partial differential equations,'' 2023. [Online]. Available: \url{https://arxiv.org/abs/2111.03794}
\BIBentrySTDinterwordspacing

\bibitem{var_pino}
\BIBentryALTinterwordspacing
M.~S. Eshaghi, C.~Anitescu, M.~Thombre, Y.~Wang, X.~Zhuang, and T.~Rabczuk, ``Variational physics-informed neural operator (vino) for solving partial differential equations,'' \emph{Computer Methods in Applied Mechanics and Engineering}, vol. 437, p. 117785, 2025. [Online]. Available: \url{https://www.sciencedirect.com/science/article/pii/S004578252500057X}
\BIBentrySTDinterwordspacing

\bibitem{phychemnode}
T.~Kumar, A.~Kumar, and P.~Pal, ``A physics-informed autoencoder-neuralode framework (phy-chemnode) for learning complex fuel combustion kinetics,'' in \emph{NeurIPS Machine Learning and the Physical Sciences Workshop}, 2024, p.~1.

\bibitem{piml_inverse_design}
\BIBentryALTinterwordspacing
S.~Sarkar, A.~Ji, Z.~Jermain, R.~Lipton, M.~Brongersma, K.~Dayal, and H.~Y. Noh, ``Physics‐informed machine learning for inverse design of optical metamaterials,'' \emph{Advanced Photonics Research}, vol.~4, no.~12, 10 2023. [Online]. Available: \url{https://www.osti.gov/biblio/2067627}
\BIBentrySTDinterwordspacing

\bibitem{piml_inverse_design_2}
\BIBentryALTinterwordspacing
L.~Lu, R.~Pestourie, W.~Yao, Z.~Wang, F.~Verdugo, and S.~G. Johnson, ``Physics-informed neural networks with hard constraints for inverse design,'' \emph{SIAM Journal on Scientific Computing}, vol.~43, no.~6, pp. B1105--B1132, 2021. [Online]. Available: \url{https://doi.org/10.1137/21M1397908}
\BIBentrySTDinterwordspacing

\bibitem{piml_inverse_design_3}
\BIBentryALTinterwordspacing
Z.~Hao, S.~Liu, Y.~Zhang, C.~Ying, Y.~Feng, H.~Su, and J.~Zhu, ``Physics-informed machine learning: A survey on problems, methods and applications,'' 2023. [Online]. Available: \url{https://arxiv.org/abs/2211.08064}
\BIBentrySTDinterwordspacing

\bibitem{neural_simulator_1}
\BIBentryALTinterwordspacing
J.~Donnelly, A.~Daneshkhah, and S.~Abolfathi, ``Physics-informed neural networks as surrogate models of hydrodynamic simulators,'' \emph{Science of The Total Environment}, vol. 912, p. 168814, 2024. [Online]. Available: \url{https://www.sciencedirect.com/science/article/pii/S0048969723074430}
\BIBentrySTDinterwordspacing

\bibitem{stochastic_de}
\BIBentryALTinterwordspacing
J.~Jia and A.~R. Benson, ``Neural jump stochastic differential equations,'' 2020. [Online]. Available: \url{https://arxiv.org/abs/1905.10403}
\BIBentrySTDinterwordspacing

\bibitem{rough_de}
J.~Morrill, C.~Salvi, P.~Kidger, J.~Foster, and T.~Lyons, ``Neural rough differential equations for long time series,'' \emph{ICML}, 2021.

\bibitem{dormand_prince}
\BIBentryALTinterwordspacing
J.~Dormand and P.~Prince, ``A family of embedded runge-kutta formulae,'' \emph{Journal of Computational and Applied Mathematics}, vol.~6, no.~1, pp. 19--26, 1980. [Online]. Available: \url{https://www.sciencedirect.com/science/article/pii/0771050X80900133}
\BIBentrySTDinterwordspacing

\bibitem{lsoda}
A.~C. Hindmarsh and L.~R. Petzold, ``Lsodar, ordinary differential equation solver for stiff or non-stiff system with root-finding,'' 2005.

\bibitem{kidger_thesis}
P.~Kidger, ``{O}n {N}eural {D}ifferential {E}quations,'' Ph.D. dissertation, University of Oxford, 2021.

\bibitem{adam_optimizer}
\BIBentryALTinterwordspacing
D.~P. Kingma and J.~Ba, ``Adam: A method for stochastic optimization,'' 2017. [Online]. Available: \url{https://arxiv.org/abs/1412.6980}
\BIBentrySTDinterwordspacing

\bibitem{cosine_annealing}
\BIBentryALTinterwordspacing
I.~Loshchilov and F.~Hutter, ``Sgdr: Stochastic gradient descent with warm restarts,'' 2017. [Online]. Available: \url{https://arxiv.org/abs/1608.03983}
\BIBentrySTDinterwordspacing

\bibitem{gru_paper}
\BIBentryALTinterwordspacing
J.~Chung, C.~Gulcehre, K.~Cho, and Y.~Bengio, ``Empirical evaluation of gated recurrent neural networks on sequence modeling,'' 2014. [Online]. Available: \url{https://arxiv.org/abs/1412.3555}
\BIBentrySTDinterwordspacing

\bibitem{lstm_paper}
S.~Hochreiter and J.~Schmidhuber, ``Long short-term memory,'' \emph{Neural Computation}, vol.~9, no.~8, pp. 1735--1780, 1997.

\bibitem{tcn_paper}
\BIBentryALTinterwordspacing
C.~Lea, R.~Vidal, A.~Reiter, and G.~D. Hager, ``Temporal convolutional networks: A unified approach to action segmentation,'' 2016. [Online]. Available: \url{https://arxiv.org/abs/1608.08242}
\BIBentrySTDinterwordspacing

\bibitem{transformer_paper}
\BIBentryALTinterwordspacing
A.~Vaswani, N.~Shazeer, N.~Parmar, J.~Uszkoreit, L.~Jones, A.~N. Gomez, L.~Kaiser, and I.~Polosukhin, ``Attention is all you need,'' 2023. [Online]. Available: \url{https://arxiv.org/abs/1706.03762}
\BIBentrySTDinterwordspacing

\bibitem{curriculum_learning}
\BIBentryALTinterwordspacing
Y.~Bengio, J.~Louradour, R.~Collobert, and J.~Weston, ``Curriculum learning,'' in \emph{ICML}, 2009. [Online]. Available: \url{https://doi.org/10.1145/1553374.1553380}
\BIBentrySTDinterwordspacing

\end{thebibliography}
\end{document}